\documentclass{article}

\usepackage[preprint,nonatbib]{nips_2018}
\usepackage[square,sort,comma,numbers]{natbib}

\usepackage[utf8]{inputenc} 
\usepackage[T1]{fontenc}    
\usepackage{hyperref}       
\hypersetup{
    citecolor=red,
    colorlinks=true,
    linkcolor=blue,
    filecolor=magenta,
    urlcolor=cyan,
}
\usepackage{url}            
\usepackage{booktabs}       
\usepackage{amsfonts}       
\usepackage{nicefrac}       
\usepackage{microtype}      
\usepackage{todonotes}      
\usepackage{subfig}         
\usepackage{algorithm2e}
\usepackage{bm}

\newcommand{\newterm}[1]{{\bf #1}}

\def\eqref#1{equation~\ref{#1}}

\def\1{\bm{1}}

\DeclareMathAlphabet{\mathsfit}{\encodingdefault}{\sfdefault}{m}{sl}
\SetMathAlphabet{\mathsfit}{bold}{\encodingdefault}{\sfdefault}{bx}{n}

\title{TensorFuzz: Debugging Neural Networks with Coverage-Guided Fuzzing}

\author{
  Augustus Odena\\
  Google Brain\\
  \And
  Ian Goodfellow\\
  Google Brain\\
}

\begin{document}

\maketitle

\begin{abstract}
  Machine learning models are notoriously difficult to interpret and debug.
  This is particularly true of neural networks.
  In this work, we introduce automated software testing techniques
  for neural networks that are well-suited to discovering
  errors which occur only for rare inputs.
  Specifically, we develop
	\newterm{coverage-guided fuzzing} (CGF) methods for neural networks.
  In CGF, random mutations of inputs to a neural network are guided by a
  coverage metric toward the goal of satisfying user-specified constraints.
  We describe how fast approximate nearest neighbor algorithms can provide this
  coverage metric.
  We then discuss the application of CGF to the following goals:
  finding numerical errors in trained neural networks, generating disagreements
  between neural networks and quantized versions of those networks, and
  surfacing undesirable behavior in character level language models.
  Finally, we release an open source library called TensorFuzz
  that implements the described techniques.
\end{abstract}

\section{Introduction}

Neural networks are gradually becoming used in more contexts that affect
human lives, including for medical
diagnosis \citep{gulshan2016development},
in autonomous vehicles \citep{huval2015empirical,angelova2015real,bojarski2016end},
as input into corporate and judicial decision making processes
\citep{scarborough2006neural,berk2017fairness},
in air traffic control \citep{katz2017reluplex},
and in power grid control \citep{siano2012real}.

Neural networks have the potential to transform these applications, saving
lives and offering benefits to more people than would be feasible using
human labor.
However, before this can be achieved, it is essential to ensure that neural
networks are reliable when used in these contexts.

Machine learning models are notoriously difficult to debug or interpret
\citep{lipton2016mythos} for a variety of reasons, ranging from the conceptual
difficulty of specifying what the user wishes to know about the model in
formal terms to statistical and computational difficulties in obtaining
answers to formally specified questions.
This property has arguably contributed to the recent `reproducibility crisis'
in machine learning
\citep{ke2017reproducibility,matters,fedus2018many,OBOBORG,gansequal,SOTA,SSL}
-- it's tricky to make robust experimental conclusions about techniques that are
hard to debug.
Neural networks can be particularly difficult to debug because even relatively
straightforward formal questions about them can be computationally
expensive to answer and because software implementations of neural networks
can deviate significantly from theoretical models.
For example, ReluPlex \citep{katz2017reluplex} can formally verify some
properties of neural networks but is too computationally expensive to
scale to model sizes used in practice.
Moreover, ReluPlex works by analyzing the description of a ReLU network as
a piecewise linear function, using a theoretical model in which all of the
matrix multiplication operations are truly linear.
In practice, matrix multiplication on a digital computer is not linear
due to floating point arithmetic, and machine learning algorithms can learn
to exploit this property to perform significantly nonlinear computation
\citep{foerster17nonlinear}.
This is not to criticize ReluPlex but to illustrate the need for additional
testing methodologies that interact directly with software as it actually
exists in order to correctly test even software that deviates from theoretical
models.

In this work, we leverage an existing technique from traditional software
engineering---\newterm{coverage-guided fuzzing} (CGF) \citep{AFL, LIBFUZZER}
---and adapt it to be applicable
to testing neural networks.
In particular, this work makes the following contributions:

\begin{itemize}

\item We introduce the notion of CGF for neural networks and describe how fast
approximate nearest neighbors algorithms can be used to check for coverage in
a general way.

\item We open source a software library for CGF called TensorFuzz.\footnote{
To be released upon publication.
}

\item We use TensorFuzz to find numerical issues in trained neural networks,
disagreements between neural networks and their quantized versions, and
undesirable behaviors in character level language models.

\end{itemize}

\noindent\begin{minipage}{\textwidth}
   \centering
   \begin{minipage}{.45\textwidth}
     \centering
     \includegraphics[height=6cm]{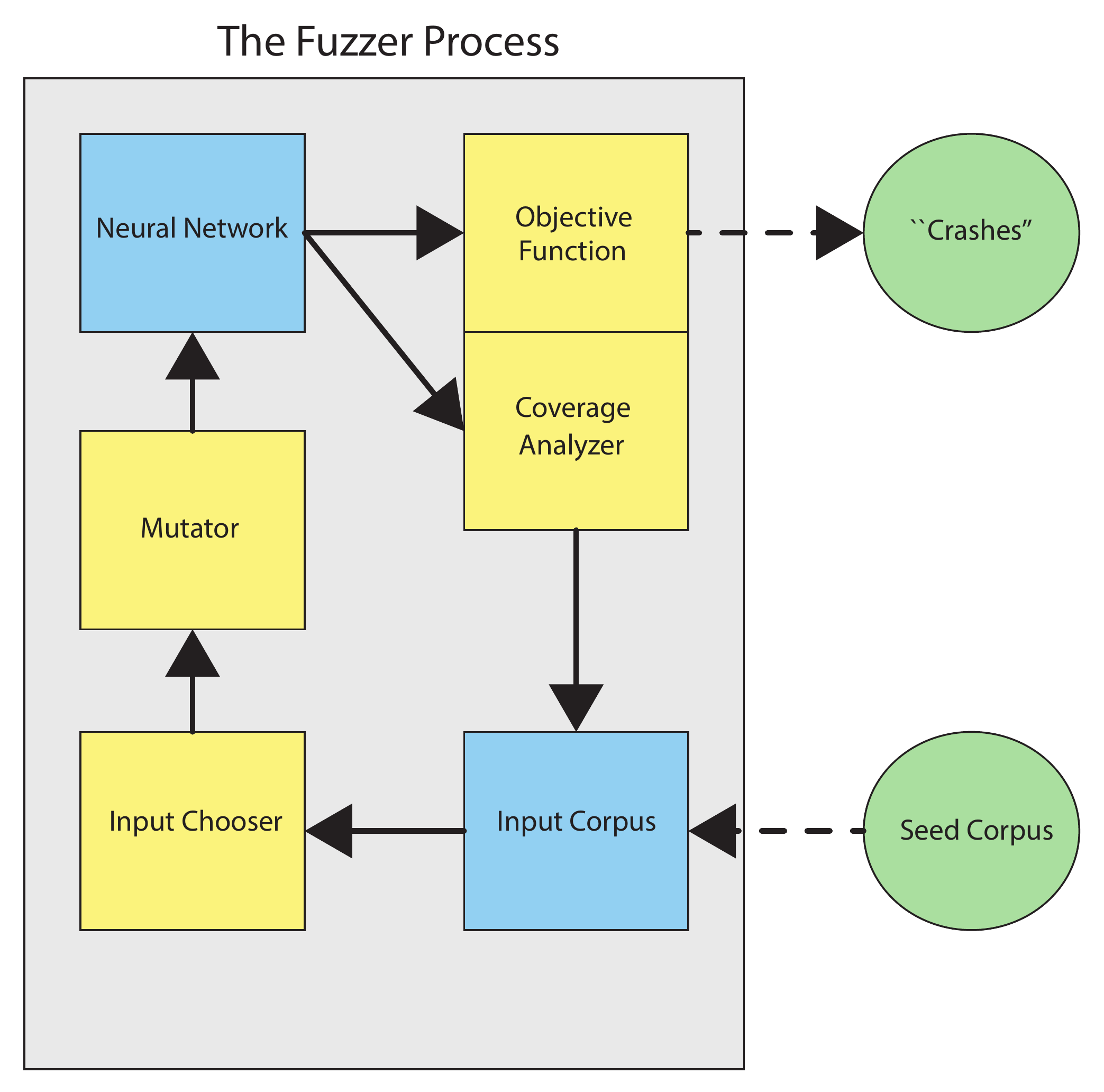}%
   \end{minipage}
   \hfill
   \begin{minipage}{.45\textwidth}
     \centering
       \begin{algorithm}[H]
        \SetAlgoLined
        \SetKwFunction{SampleFromCorpus}{SampleFromCorpus}
        \SetKwFunction{Mutate}{Mutate}
        \SetKwFunction{Fetch}{Fetch}
        \SetKwFunction{Coverage}{Coverage}
        \SetKwFunction{Metadata}{Metadata}
        \SetKwFunction{IsNewCoverage}{IsNewCoverage}
        \SetKwFunction{Objective}{Objective}
        \KwData{A seed-corpus of inputs to the computation graph}
        \KwResult{Test cases satisfying the objective}
         \While{number of iterations < N}{
          $parent$ $\leftarrow$ \SampleFromCorpus\;
          $data$ $\leftarrow$ \Mutate{$parent$}\;
          $cov, meta$ $\leftarrow$ \Fetch{$data$}\;
          \If{\IsNewCoverage{$cov$}}{
           add data to corpus\;
           }
          \If{\Objective{$meta$}}{
           add data to list of test cases\;
           }
         }
        \end{algorithm}
   \end{minipage}
   \captionof{figure}{
   Coarse descriptions of the main fuzzing loop.
   Left: A diagram of the fuzzing procedure, indicating the flow of data.
   Right: A description of the main loop of the fuzzing procedure in algorithmic
   form.
   }
   \label{fig:diagram}
\end{minipage}

\section{Background}

This section gives background on coverage-guided fuzzing for traditional
software, then discusses existing methods for testing neural networks before
finally touching on how CGF relates to these methods.

\subsection{Coverage-guided fuzzing}
\label{section:coverageguidedfuzzing}

Coverage-guided fuzzing is used to find many serious bugs in real software
\citep{OSSFUZZ}.
Two of the most popular coverage-guided fuzzers for normal
computer programs are AFL \citep{AFL} and libFuzzer \citep{LIBFUZZER}.
These have been expanded in various ways in order to make them faster or to
increase the extent to which certain parts of the code can be targeted
\citep{FUZZINGASAMARKOVCHAIN, DIRECTEDGREYBOXFUZZING}.

In coverage-guided fuzzing, a fuzzing process maintains
an input corpus containing inputs to the program under consideration.
Random changes are made to those inputs according to some mutation procedure,
and mutated inputs are kept in the corpus when they exercise new ``coverage''.
What is coverage? It depends on the type of fuzzer and on the goals at hand.
One common measure is the set of parts of the code that have been executed.
By this measure, if a new input causes the code to branch a different way at an
if-statement than it has previously, coverage has increased.

CGF has been highly successful at identifying defects in traditional
software, so it is natural to ask whether it could be applied to
neural networks.
Traditional coverage metrics track which lines of code have been executed
and which branches have been taken.
In their most basic forms, neural networks are implemented as a sequence of
matrix multiplications followed by elementwise operations.
The underlying software implementation of these operations may contain many
branching statements but many of these are based on the size of the matrix
and thus the architecture of the neural network, so the branching behavior
is mostly independent of specific values of the neural network's input.
A neural network run on several different inputs will thus often execute the
same lines of code and take the same branches, yet produce interesting
variations in behavior due to changes in input and output values.
Executing an existing CGF tool such as AFL thus may not find interesting
behaviors of the neural network.
In this work, we elect to use fast approximate nearest neighbor algorithms to
determine if two sets of neural network `activations' are meaningfully different
from each other.
This provides a coverage metric producing useful results for neural
networks, even when the underlying software implementation of the neural
network does not make use of many data-dependent branches.

\subsection{Testing of Neural Networks}
\label{section:testingofneuralnetworks}

Methods for testing and computing test coverage of traditional computer programs
cannot be straightforwardly applied to neural networks.
We can't just naively compute branch coverage, for example, because of reasons
discussed above.
Thus, we have to think about how to write down useful coverage metrics for
neural networks.
Though this work is the first (as far as we know) to explore the idea of CGF
for neural networks, it's not the first to address the issues of testing and
test coverage for neural networks.
A variety of proposals (many of which focus on adversarial
examples \citep{ADVX}) have been made for ways to test
neural networks and to measure their test coverage.
We survey these proposals here:

\citet{DEEPXPLORE} introduce the metric of neuron coverage for a neural network
with rectified linear units (ReLus) as the activation functions.
A test suite is said to achieve full coverage under this metric if for every
hidden unit in the neural network, there is some input for which that hidden
unit has positive value. They then cross reference multiple neural
networks using gradient based optimization to find misbehavior.

\citet{DEEPGAUGE} generalize neuron coverage in two ways.
In k-multisection coverage, they take -- for each neuron -- the range of values
seen during training, divide it into k chunks, and measure whether each of the k
chunks has been ``touched''.
In neuron boundary coverage, they measure whether each activation has been made
to go above and below a certain bound.
They then evaluate how well these metrics are satisfied by the test set.

\citet{TESTINGDEEPNEURALNETWORKS} introduce a family of metrics inspired by
Modified Condition / Decision Coverage \citep{MCDC}.
We describe their ss-coverage proposal as an example:
Given a neural network arranged into layers, a pair of neurons ($n_1, n_2$) in
adjacent layers is said to be ss-covered by a pair $(x, y)$ of inputs if the
following 3 things are true: $n_1$ has a different sign for each of $x, y$,
$n_2$ has a different sign for each of $x, y$, and all other elements of the
layer containing $n_1$ have the same sign for $x, y$.

\citet{DEEPTEST} applies the neuron coverage metric to deep neural networks that
are part of self-driving car software. They perform natural image
transformations such as blurring, shearing, and transformation and use the
idea of metamorphic testing \citep{METAMORPHIC} to find errors.

\citet{BLACKBOX} perform black box testing of image classifiers using
image-specific operations.
Concurrent with this work, \citet{CONCOLIC} leverages a complementary approach
called concolic execution. Whereas our approach is analogous to AFL or
libFuzzer, their approach is analogous to CUTE \citep{CUTE}.

\subsection{Opportunities for improvement}
\label{section:opportunitiesforimprovement}

It is heartening that so much progress has been made recently on the problem of
testing neural networks. However, the success of e.g. AFL and libFuzzer in spite
of the existence of more sophisticated techniques suggests that there is a role
for an analogous tool that works on neural networks.
Ideally we would implement CGF for neural networks using the coverage metrics
from above. However, all of these metrics, though perhaps appropriate in the
context originally proposed, lack certain desirable qualities.
We describe below why this is true for the most relevant
metrics.

\citet{TESTINGDEEPNEURALNETWORKS} claim that the neuron coverage
metric is too easy to satisfy. In particular, they show that 25
randomly selected images from the MNIST test set yield close to 100\% neuron
coverage for an MNIST classifier. This metric is also specialized to work on
rectified linear units (reLus) which limits its generality.

Neuron boundary coverage \citep{DEEPGAUGE} is nice in that it doesn't rely on
using reLus, but it also still treats neurons independently.
This causes it to suffer from the same problem as neuron
coverage: it's easy to exercise all of the coverage with few examples.

The metrics from \citet{TESTINGDEEPNEURALNETWORKS} are an improvement on neuron
coverage and may be useful in the context of more formal methods,
but for our desired application, they have several disadvantages.
They still treat reLus as a special case,
they require special
modification to work with convolutional neural networks,
and they do not offer an obvious generalization
that supports
attention \citep{ATTENTION} or residual networks \citep{RESNETS}.
They also rely on neural networks being arranged in hierarchical layers, which
is often not true for modern deep learning architectures.

What we would really like is a coverage metric that is simple, cheap to compute,
and is easily applied to all sorts of neural network architectures.
Thus, we propose storing the activations (or some subset of them) associated
with each input, and checking whether coverage has increased on a given input
by using an approximate nearest neighbors algorithm to see whether there are any
other sets of activations within a pre-specified distance.
We discuss this idea in more detail in Section \ref{section:coverageanalyzer}.

\section{The TensorFuzz library}

Drawing inspiration from the fuzzers described in the previous section, we have
implemented a tool that we call TensorFuzz.
It works in a way that is analogous to those tools, but that is different in
ways that make it more suitable for neural network testing.
Instead of an arbitrary computer program written in \textbf{C} or \textbf{C++},
it feeds inputs to an arbitrary TensorFlow graph.
Instead of measuring coverage by looking at basic blocks or changes in control
flow, it measures coverage by (roughly speaking) looking at the ``activations''
of the computation graph.
In \ref{section:basicfuzzingprocedure}, we discuss the overall architecture of
the fuzzer, including the flow of data and the basic building blocks.
In \ref{section:fuzzingproceduredetails}, we discuss building blocks in more
detail. In particular, we describe how the corpus is sampled from, how mutations
are performed, and how coverage and objective functions are evaluated.
We pay extra attention to describing special
challenges associated with defining a coverage metric for neural networks and
explain how these challenges can be (at least partially) dealt with using
approximate nearest neighbors.

\subsection{The basic fuzzing procecure}
\label{section:basicfuzzingprocedure}
The overall structure of the fuzzing procedure is very similar to the structure
of coverage-guided fuzzers for normal computer programs.
The main difference is that instead of interacting with an arbitary computer
program that we have instrumented, we interact with a TensorFlow graph that we
can feed inputs to and get outputs from.

The fuzzer starts with a seed corpus containing at least one set of inputs for
the computation graph. Unlike in traditional CGF, we don't just feed in big
arrays of bytes. Instead, we restrict the inputs to those that are in some sense
valid neural network inputs. If the inputs are images, we restrict our inputs to
have the correct size and shape, and to lie in the same interval as the input
pixels of the dataset under consideration. If the inputs are sequences of
characters, we only allow characters that are in the vocabulary extracted from
the training set.

Given this seed corpus, fuzzing proceeds as follows:
Until instructed to stop, the fuzzer chooses elements from the input corpus
according to some component we will call the Input Chooser.
For the purpose of this section, it's ok to imagine the Input Chooser as
choosing uniformly at random, though we will describe more complicated
strategies in \ref{section:fuzzingproceduredetails}.

Given an input, the Mutator component will perform some sort of modification to
that input. The modification can be as simple as just flipping the sign of an
input pixel in an image, and it can also be restricted to follow some kind of
constraint on the total modification made to a corpus element over time -
see \ref{section:fuzzingproceduredetails} for more on this.

Finally, the mutated inputs can be fed to the neural network.
In TensorFuzz, two things are extracted from the neural network:
a set of coverage arrays, from which the actual coverage will be computed,
and a set of metadata arrays, from which the result of the objective function
will be computed.

Once the coverage is computed, the mutated input will be added to the corpus if
it exercises new coverage, and it will be added to the list of test cases if it
causes the objective function the be satisfied.

See Figure \ref{fig:diagram} for complementary overviews of this procedure.

\subsection{Details of the fuzzing procedure}
\label{section:fuzzingproceduredetails}

In this section we describe in more detail the components of the fuzzer.

\paragraph{Input Chooser:}
At any given time, the fuzzer must choose which inputs from the existing corpus
to mutate. The optimal choice will of course be problem dependent, and
traditional CGFs rely on a variety of heuristics to make this determination.
For the applications we tested, making a uniform random selection worked
acceptably well, but we ultimately settled on the following heuristic, which we
found to be faster:
\begin{math}
p(c_k, t) = \frac{e^{t_k - t}}{\sum e^{t_k - t}}
\end{math},
where $p(c_k, t)$ gives the probability of choosing corpus element $c_k$ at time
$t$ and $t_k$ is the time when element $c_k$ was added to the corpus. The
intuition behind this is that recently sampled inputs are more likely to yield
useful new coverage when mutated, but that this advantage decays as time
progresses.

\paragraph{Mutator:}
Once the Input Chooser has chosen an element of the corpus to mutate, the
mutations need to be applied. In this work, we had to implement mutations for
both image inputs and text inputs.
For image inputs, we implemented two different types
of mutation. The first is to just add white noise of a user-configurable
variance to the input.
The second is to add white noise, but to constrain the difference between the
mutated element and the original element from which it is descended to have a
user-configurable $L_{\infty}$ norm.
This type of constrained mutation can be useful if we want to find inputs that
satisfy some objective function, but are still plausibly of the same ``class''
as the original input that was used as a seed.
In both types of image mutation we clip the image after mutation so that it
lies in the same range as the inputs used to train the neural network being
fuzzed.

For text inputs, since we can't simply add uniform noise to
the string, we mutate according to the following policy:
we uniformly at random perform one of these operations:
delete a character at a random location,
add a random character at a random location,
or substitute a random character at a random location.

\paragraph{Objective Function:}
Generally we will be running the fuzzer with some goal in mind.
That is, we will want the neural network to reach some particular state - maybe
a state that we regard as erroneous.
The objective function is used to assess whether that state has been
reached.
When the mutated inputs are fed into the computation graph, both coverage arrays
and metadata arrays are returned as output.
The objective function is applied to the metadata arrays, and flags inputs that
caused the objective to be satisfied.

\paragraph{Coverage Analyzer:}
\label{section:coverageanalyzer}
The coverage analyzer is in charge of reading arrays from the TensorFlow
runtime, turning them into python objects representing coverage, and checking
whether that coverage is new.
The algorithm by which new coverage is checked is central to the proper
functioning of the fuzzer.

The characteristics of a desirable coverage checker are: we want it to check if
the neural network is in a `state' that it hasn't been in before, so that
we can find misbehaviors that might not be caught by the test set.
We want this check to be fast (so we probably want it to be simple), so that
we can find many of those misbehaviors quickly.
We want it to work for many different types of computation graph without special
engineering, so that practitioners can use our tooling without having to make
special adaptations.
We want exercising all of the coverage to be hard, otherwise we won't actually
cover very much of the possible behaviors.
Finally, we want getting new coverage to help us make incremental progress, so
that continued fuzzing yields continued gains.

As alluded to in Section \ref{section:opportunitiesforimprovement}, none the
coverage metrics discussed in Section \ref{section:testingofneuralnetworks}
quite meet all these desiderata,
but we can design from first principles a coverage metric that at least
comes closer to meeting them.

A naive, brute force solution is to read out the whole activation
vector and treat new activation vectors as new coverage.
However, such a coverage metric would not provide useful guidance,
because most inputs would trivially yield new coverage.
It is better to detect whether an activation vector is close
to one that was observed previously.
One way to achieve this is to use an approximate nearest neighbors algorithm,
(as used for Neural Turing Machine \citep{NTM} memory by \citet{SAM}).
When we get a new activation vector, we can look up its nearest neighbor, then
check how far away the nearest neighbor is in Euclidean distance and add the
input to the corpus if the distance is greater than some amount $L$.

This is essentially what we do.
Currently, we use an open source libarary called FLANN \citep{FLANN} to compute
the approximate nearest neighbors.
In general, you may not need to see all the activations.
The set of activations to use is presently a matter for empirical tuning.
We find it is often possible to obtain good results by tracking only
the logits, or the layer before the logits.

One potential optimization to note:
you don't actually need to know the nearest neighbor, you just need to
know whether there exists a neighbor within some range.
For this, you can use a distance-sensitive Bloom filter \citep{
DISTANCESENSITIVEBLOOMFILTERS}, but this will come at the expense of sometimes
counting coverage as `not new' when it is new.
If your fuzzing application is one in which this is acceptable, this could speed
things up for you. However, we leave the testing of this idea to future work.

\subsection{Batching and nondeterminism}
\label{section:batching}

There are two other issues that are somewhat unique to TensorFlow graphs that
merit a few words.
First, nearly all TensorFlow graphs that exist in practice have been engineered
to take advantage of hardware parallelism offered by e.g. modern GPUs\footnote{
In fact, they've often been engineered to run on multiple GPUs, which suggests
an obvious optimization to TensorFuzz: unlike stochastic gradient descent,
in which parallelism is limited by stale gradients \citep{FASGD}, fuzzing
performance can theoretically scale linearly with hardware, assuming that the
corpus does not need to be kept perfectly synchronized.}
For this reason, it could be wasteful to only fetch the coverage and metadata
for one mutated input at a time.
Instead, we perform the mutations as a batch and feed a batch of inputs to the
computation graph, then we check the coverage and objective function on a batch
of output arrays.
Second, computation graphs will often give nondeterministic outputs - both
because of instructions that execute nondeterministically (e.g. large
accumulations on GPUs) and because of fundamentally random operations built into
the graph. For now, we have chosen to deal with this in the most naive way
possible: if we can feed the same input in twice and get different coverage, we
simply include the input twice in the corpus.

\section{Experimental results}
\label{section:experiments}
We briefly present a few different applications of the CGF technique to
establish that it is useful in general settings.

\subsection{CGF can efficiently find numerical errors in trained neural
networks}
\label{section:nans}

Since neural networks use floating point math \citep{FLOATINGPOINT}, they are
susceptible to numerical issues, both during training and at evaluation time.
These issues are notoriously hard to debug, partly because they may only be
triggered by a small set of rarely encountered inputs.
This is one case where CGF can help.
We focus on finding inputs that result in not-a-number (NaN) values.

\paragraph{Numerical errors are important to find:}

Numerical errors, especially those resulting in NaNs, could cause dangerous
behavior of important systems if these errors are first encountered
`in the wild`.
CGF can be used to identify a large number of errors before deployment,
and reduce the risk of errors occurring in a harmful setting.

\begin{figure}%
\begin{center}
\resizebox{.9\textwidth}{!}{%
\includegraphics[height=3cm]{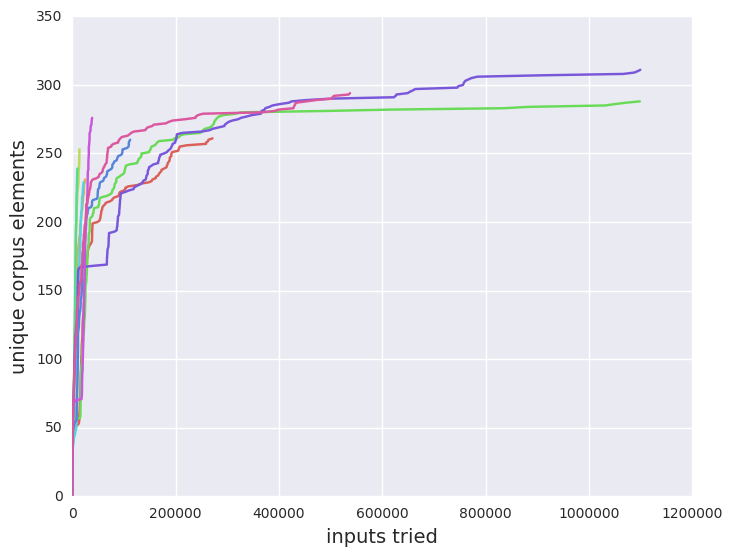}%
\quad
\includegraphics[height=3cm]{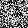}%
}
\caption{
We trained an MNIST classifier with some unsafe numerical operations.
We then ran the fuzzer 10 times on random seeds from the MNIST
dataset.
The fuzzer found a non-finite element every run.
Random search never found a non-finite element.
Left: the accumulated corpus size of the fuzzer while running, for 10 runs.
Right: an example satisfying image found by the fuzzer.
}
\label{fig:nanfigure}
\end{center}
\end{figure}

\paragraph{CGF can quickly find numerical errors:}
With CGF, we should be able to simply add check numerics ops to the metadata
and run our fuzzer. To test this hypothesis, we trained a fully connected neural
network to classify MNIST \citep{MNIST} digits.
We used a poorly implemented cross entropy loss on purpose so that there would
be some chance of numerical errors.
We trained the model for 35000 steps with a mini-batch size of 100, at which
point it had a validation accuracy of 98\%.
We then checked that there were no elements in the MNIST dataset that
cause a numerical error.
Nevertheless, TensorFuzz found NaNs quickly across multiple random
initializations.
See Figure \ref{fig:nanfigure} for more details.

\paragraph{Gradient-based search techniques might not help find numerical errors:}
One potential objection to CGF techniques is that gradient-based search
techniques might be more efficient than more randomized search techniques.
However, it is not obvious how to specify the objective for a gradient based
search.
There is not a straightforward way to measure how similar a real-valued
output of the model is to a NaN value.

\paragraph{Random search is prohibitively inefficient for finding some numerical errors:}
To establish that random search is not sufficient and that coverage-guidance
is necessary for efficiency, we compared to random search.
We implemented a baseline random search algorithm and ran it for 100,000 samples
from the corpus with 10 different random initializations.
The baseline was not able to find a non-finite element in any of these trials.

\subsection{CGF surfaces disagreements between models and their quantized
versions}

Quantization \cite{QNN} is a process by which neural network weights are stored
and neural network computations performed using numerical representations that
consist of fewer bits of computer memory.
Quantization is a popular approach to reducing the computational cost
or size of neural networks, and is widely used for e.g. running inference on
cell phones as in \href{
https://developer.android.com/ndk/guides/neuralnetworks/}{
Android Neural Networks API} or  \href{
https://www.tensorflow.org/mobile/tflite/}{TFLite} and in the context of custom
machine learning hardware -- e.g. Google's Tensor Processing Unit \citep{TPU} or
\href{https://developer.nvidia.com/tensorrt}{NVidia's TensorRT}.

\paragraph{Errors resulting from quantization are important to find:}
Of course, quantization is not very useful if it reduces the accuracy of the
model too dramatically. Given a quantized model, it would thus be nice to
check how much quantization reduced the accuracy.

\paragraph{Few errors can be found just by checking existing data:}
As a baseline experiment, we trained an MNIST classifier (this time without
intentionally introducing numerical issues) using 32-bit floating point numbers.
We then truncated all weights and activations to 16-bits.
We then compared the predictions of the 32-bit and the 16-bit model on the MNIST
test set and found 0 disagreements.

\begin{figure}
\begin{center}
\resizebox{.9\textwidth}{!}{%
\includegraphics[height=3cm]{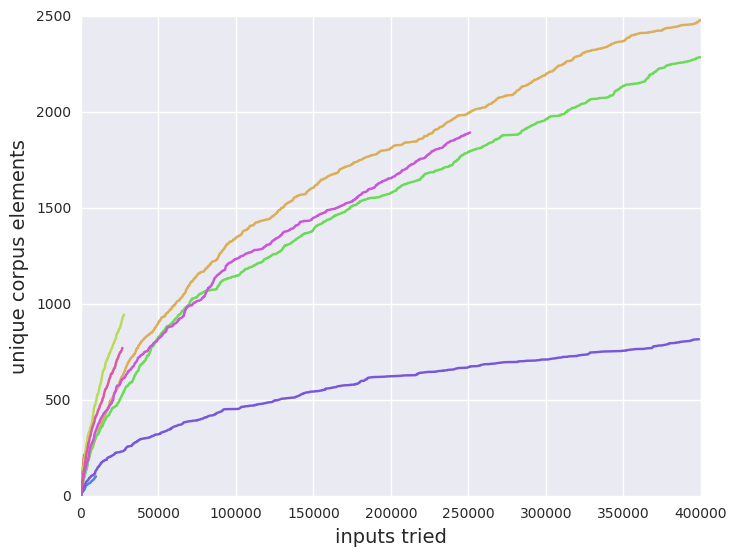}%
\quad
\includegraphics[height=3cm]{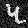}%
}
\caption{
We trained an MNIST classifier with 32-bit floats and then truncated the
associated TensorFlow graph to 16-bit floats.
Both the original and the truncated graph made the same predictions on all 10000
elements of the MNIST test set, but the fuzzer was able to find disagreements
within an infinity-norm ball of radius 0.4 around 70\% of the test images that
we tried to fuzz.
Left: the accumulated corpus size of the fuzzer while running, for 10 runs.
Lines that go all the way to the right correspond to failed fuzzing runs.
Right: an image found by the fuzzer that is classified differently by the 32-bit
and 16-bit neural networks.
}
\label{fig:quantizefigure}
\end{center}
\end{figure}

\paragraph{CGF can quickly find many errors in small regions around the data:}
We then ran the fuzzer with mutations restricted to lie in a radius 0.4 infinity
norm ball surrounding the seed images, using the activations of only the 32-bit
model as coverage.
We restrict to inputs near seed images because these inputs nearly all
have unambiguous class semantics.
It is less interesting if two versions of the model disagree on
out-of-domain garbage data with no true class.
With these settings, the fuzzer was able to generate disagreements for 70\% of
the examples we tried.
Thus, CGF allowed us to find real errors that could have occured at test time.
See Figure \ref{fig:quantizefigure} for more details.

\paragraph{Random search fails to find new errors given the same number of
mutations as CGF:}
As in Section \ref{section:nans} we tried a baseline random search method in
order to demonstrate that the coverage-guidance specifically was useful in this
context. The random search baseline was not able to find any disagreements that
didn't already exist when given the same number of mutations as the fuzzer.

\subsection{CGF surfaces undesirable behavior in character level language
models}
\label{section:rnns}

Given a trained machine learning model, we may be able to characterize certain
of its behaviors as undesirable, even though incorporating this characterization
into the training loss might be inconvenient or even intractable.
As an example, we train a character level
language model of the type described in \citet{UNREASONABLE}.
In particular, we modify the code from \href{
https://github.com/sherjilozair/char-rnn-tensorflow}{
https://github.com/sherjilozair/char-rnn-tensorflow} to train a 2 layer LSTM
\citep{LSTM} on the Tiny Shakespeare \citep{VISUALIZING} dataset.

We then consider the application of sampling from this trained language model
given a priming string. One can imagine such a thing being done in an
auto-complete application, for example. For illustrative purposes, we identify
two desiderata that we can approximately enforce via the fuzzer: First, the
model should not repeat the same word too many times in a row. Second, the model
should not output words from the blacklist.

We fuzz the model using for coverage the hidden state of the LSTM after the
priming string has been consumed. We sample from the model according to a
softmax over the logits, using a fixed random seed that we reset at every
sampling.
We use the mutation function described in Section
\ref{section:fuzzingproceduredetails}.

We ran an instance of TensorFuzz and an instance of random search for 24 hours
each. TensorFuzz was able to generate repeat words, but so was random search.
However, TensorFuzz was able to generate six out of ten words from our blacklist
while random search only generated one.

\section{Conclusion}
We have introduced the concept of coverage-guided fuzzing for neural networks
and described how to build a useful coverage checker in this context.
We have demonstrated the practical applicability of TensorFuzz by finding
numerical errors, finding disagreements between neural networks and their
quantized versions, and surfacing undesirable behavior in RNNs.
Finally, we are releasing along with this paper the implementation of
TensorFuzz, so that other people can both build on our work and also use our
fuzzer to find real issues.

\subsubsection*{Acknowledgments}
We thank Dave Anderson for helpful comments on an early version of the draft.
We thank Rishabh Singh, Alexey Kurakin, and Martín Abadi for general input.
We also thank Kostya Serebryany for helpful explanations of libFuzzer.

\clearpage
\clearpage

\bibliography{paper}
\bibliographystyle{plainnat}

\end{document}